\documentclass[11pt]{article}
\usepackage[utf8]{inputenc}
\usepackage{textgreek}
\usepackage{comment}

\usepackage[preprint]{acl}
\usepackage{xcolor}

\usepackage{times}
\usepackage{latexsym}
\usepackage{amsmath}
\usepackage{placeins}  % Provides \FloatBarrier
\usepackage{booktabs}
\usepackage{multirow}
\usepackage{graphicx}
\usepackage{algorithm}
\usepackage{algpseudocode}
\usepackage{tcolorbox}
\usepackage[T1]{fontenc}
\usepackage[utf8]{inputenc}
\usepackage{amssymb}
\usepackage{tcolorbox}

\newtcolorbox{promptbox}[1][]{
  colback=white,
  colframe=gray!60,
  coltitle=white,
  colbacktitle=gray!60,
  fonttitle=\bfseries\large,
  title={Prompt},
  rounded corners,
  boxrule=0.5pt,
  arc=3mm,
  #1
}

\newtcolorbox{custompromptbox}[2][]{
  colback=white,
  colframe=gray!60,
  coltitle=white,
  colbacktitle=gray!60,
  fonttitle=\bfseries\large,
  title={#2},
  rounded corners,
  boxrule=0.5pt,
  arc=3mm,
  #1
}

\usepackage{microtype}

\usepackage{tikz}
\usetikzlibrary{arrows.meta, positioning, shapes.geometric}

\usepackage{inconsolata}

\usepackage{graphicx}

\title{Facet-Level Tracing of Evidence Uncertainty and Hallucination in RAG}

\author{
Passant Elchafei$^1$, \
Monorama Swain$^1$, \
Shahed Masoudian$^1$, \
Markus Schedl$^{1,2}$ \\
$^1$Johannes Kepler University Linz, Institute of Computational Perception, Linz, Austria \\
$^2$Linz Institute of Technology, Artificial Intelligence Lab, Austria \\
\texttt{\{passant.elchafei, monorama.swain, shahed.masoudian, markus.schedl\}@jku.at}
}

\begin{document}
\maketitle
\begin{abstract}
Retrieval-Augmented Generation (RAG) aims to reduce hallucination by grounding answers in retrieved evidence, yet hallucinated answers remain common even when relevant documents are available. Existing evaluations focus on answer or passage-level accuracy, offering limited insight into how evidence is used during generation. In this work, we introduce a facet-level diagnostics framework for QA that decomposes each input question into atomic reasoning facets. For each facet, we assess evidence sufficiency and grounding using a structured Facet×Chunk matrix that combines retrieval relevance with natural language inference–based faithfulness scores. To diagnose evidence usage, we analyze three controlled inference modes: Strict RAG which enforces exclusive reliance on retrieved evidence, Soft RAG which allows integration of retrieved evidence and parametric knowledge, and LLM-only generation without retrieval. Comparing these modes enables thorough analysis of retrieval–generation misalignment, defined as cases where relevant evidence is retrieved but not correctly integrated during generation. Across medical QA and HotpotQA, we evaluate three open- and closed-source LLMs (GPT, Gemini, and LLaMA), providing interpretable diagnostics that reveal recurring facet-level failure modes, including evidence absence, evidence misalignment, and prior-driven overrides. Our results demonstrate that hallucinations in RAG systems are driven less by retrieval accuracy and more by how retrieved evidence is integrated during generation, with facet-level analysis exposing systematic evidence override and misalignment patterns that remain hidden under answer-level evaluation.
\end{abstract}
\section{Introduction}
Retrieval-Augmented Generation (RAG) is widely used to improve the factual reliability of LLM by grounding generation in external evidence \cite{Lewis2020RetrievalAugmentedGF, Guu2020REALMRL}. By conditioning answers on retrieved documents, RAG aims to reduce hallucinations, particularly in knowledge-intensive tasks such as question answering. Nevertheless, hallucinated or unsupported content remains common, even when seemingly relevant evidence is retrieved \cite{Min2023FActScoreFA, Li2023HaluEvalAL}.
\begin{figure*}[t!]
  \centering
  \includegraphics[
  width=0.85\textwidth,
  height=0.25\textheight,
  keepaspectratio
    ]{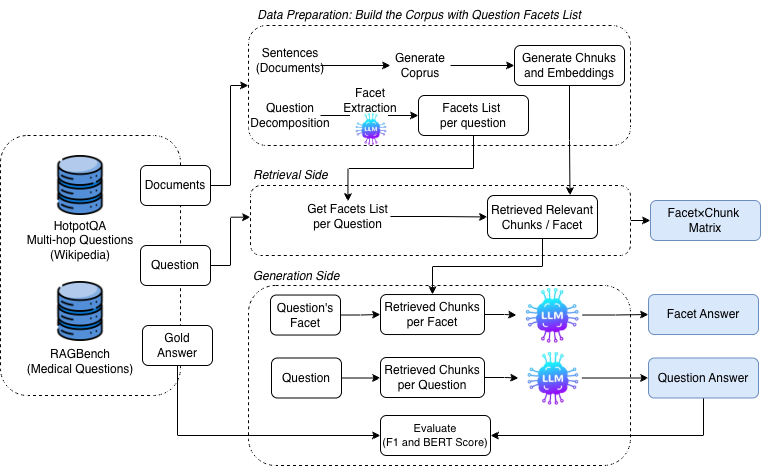}
  \caption{Facet-RAG Pipeline. The framework decomposes questions into reasoning facets, retrieves evidence at the chunk level, and evaluates facet-level entailment and generation faithfulness to diagnose hallucination and uncertainty in RAG systems.}
  \label{fig:uq_pipeline}
\end{figure*}
A prevailing assumption in RAG is that stronger retrieval leads to more faithful generation. Accordingly, prior work has focused on improving retrieval quality and evaluating systems using retrieval-centric metrics or answer-level accuracy \cite{Asai2023SelfRAGLT}. In practice, however, retrieval confidence alone often fails to guarantee faithful evidence usage, especially for complex reasoning questions \cite{wang2023context}. Models may retrieve highly relevant passages yet ignore them, partially use them, or override them with parametric knowledge—behaviors that are difficult to diagnose using standard evaluation methods \cite{chen2022rich}. Yet this relationship remains poorly understood \cite{Bohnet2022AttributedQA, cuconasu2024power}.

\textbf{\emph{Does improving retrieval quality actually result in more faithful answers in RAG systems?} } 
Our empirical analysis suggests that the answer is often \emph{No}. Across models, datasets, and difficulty levels, we find that high-quality retrieval frequently co-occurs with unfaithful generation, while faithful answers may arise even under imperfect retrieval. This indicates that hallucination in RAG is not solely a consequence of retrieval failure, but often arises from \emph{systematic retrieval--generation misalignment}. To explain this phenomenon, we introduce a facet-level diagnostic framework that decomposes each question into atomic reasoning facets. For each facet, we analyze evidence support using a structured Facet$\times$Chunk evidence matrix that combines retrieval relevance with natural language inference (NLI)–based faithfulness signals. This enables fine-grained, evidence-centric signals of model behavior that are invisible at the answer level. We further analyze three controlled inference modes—\emph{Strict RAG}, \emph{Soft RAG}, and \emph{LLM-only} generation—which we treat as diagnostic interventions to disentangle retrieval-related failures from generation-side hallucinations. While our primary contribution is diagnostic, we also examine whether facet-level evidence behaviors translate into end-to-end answer quality by aggregating results at the question level.

We evaluate our framework on multi-hop question answering using HotpotQA~\cite{yang2018hotpotqa} and on a medical question answering dataset \cite{Friel2024RAGBenchEB} derived from clinical guidelines, using three LLMs: GPT-4o mini,\footnote{GPT-4o mini is accessed via the OpenAI API.} LLaMA-3-8B-Instruct,\footnote{LLaMA-3-8B-Instruct is an open-weight model released by Meta.} and Gemini-2.0-Flash.\footnote{Gemini-2.0-Flash is accessed via the Google Generative AI API.} Our contributions include: (1) a facet-level diagnostic framework that decomposes questions into atomic reasoning steps and constructs structured evidence matrices to trace retrieval-generation interactions, (2) a fine-grained evidence taxonomy that categorizes five distinct failure modes including evidence absence, misalignment, and prior-driven overrides, and (3) controlled evaluation across these domains and models, revealing that hallucinations primarily arise from generation-side evidence misalignment rather than retrieval failure.

\begin{table*}[t]
\centering
\footnotesize
\setlength{\tabcolsep}{3pt}
\begin{tabular}{p{5cm}*{11}{c}}
\toprule
\textbf{Facet} & C1 & C2 & C3 & C4 & C5 & C6 & C7 & C8 & C9 & C10 & \textbf{Max} \\
\midrule
1) What is the type of place associated with Regency Road, Adelaide?
& 0.00 & 0.14 & 0.00 & 0.27 & 0.18 & 0.04 & 0.06 & 0.01 & 0.02 & \textbf{0.47} & \textbf{0.47} \\
2) What is the type of place associated with Klemzig, South Australia?
& 0.14 & 0.01 & 0.00 & 0.21 & 0.00 & 0.14 & 0.01 & \textbf{0.30} & 0.14 & 0.03 & \textbf{0.30} \\
\bottomrule
\end{tabular}
\caption{HotpotQA Example: Facet-level coverage over retrieved chunks (C1–C10). Main question: \emph{``What type of place does Regency Road, Adelaide and Klemzig, South Australia have in common?''} Question type: \emph{bridge}. Difficulty level: \emph{hard}.}
\label{tab:facet_coverage_bridge}
\end{table*}

\section{Related work}
RAG systems combine neural retrieval with conditional generation to ground model outputs in external knowledge \cite{Lewis2020RetrievalAugmentedGF, Guu2020REALMRL}. Prior work has focused on improving retrieval quality through dense retrievers \cite{Karpukhin2020DensePR} and iterative strategies \cite{jiang2023active, trivedi2023interleaving}. While adaptive retrieval methods \cite{Shi2023REPLUGRB} dynamically adjust retrieval behavior, they do not explicitly analyze why models may fail to utilize retrieved evidence even when it is relevant and sufficient.

Hallucination detection has been studied at the answer level \cite{Maynez2020OnFA, Li2023HaluEvalAL}, with atomic fact decomposition \cite{Min2023FActScoreFA} and self-consistency techniques \cite{Manakul2023SelfCheckGPTZB} for evaluation. Multi-hop question answering work explores chain-of-thought prompting \cite{yao2023tree} and decomposition-based reasoning \cite{press2023measuring} to improve accuracy. 

Uncertainty estimation in RAG has focused on confidence calibration \cite{xiong2023can}, semantic uncertainty \cite{kuhn2023semantic} study semantic uncertainty through token-level entropy measures, while \cite{lin2023generating} propose uncertainty-aware decoding strategies. These approaches operate at the generation level and do not identify whether uncertainty arises from retrieval ambiguity or generation-side evidence misalignment. Our work differs by introducing a facet-level diagnostic framework that causally probes evidence usage through controlled inference modes, enabling fine-grained analysis of retrieval–generation misalignment across semantic facet types.

\section{Methodology}
Our goal is to diagnose evidence uncertainty and hallucination behavior in RAG systems. Rather than proposing new retrievers or generators, our methodology: We analyze how retrieval and generation interact under controlled inference conditions to expose systematic retrieval–generation misalignment as shown in Figure~\ref{fig:uq_pipeline}.
Given a question and its retrieved documents, we decompose the question into atomic reasoning facets, evaluate evidence support for each facet, and analyze how evidence usage changes across inference modes. An illustrative facet-level diagnostic example demonstrating the full pipeline on a concrete instance is provided in Appendix~\ref{app:complete_example}.

\subsection{Datasets}
We evaluate our framework on two complementary question answering datasets designed to probe different aspects of evidence usage in RAG.
\subsubsection{HotpotQA}
HotpotQA \cite{yang2018hotpotqa} is a multi-hop question answering benchmark constructed from Wikipedia. We use a stratified subset of 3,000 questions across three difficulty levels (easy, medium, hard) to analyze evidence integration under increasing reasoning complexity.

\subsubsection{RAGBench--Medical}
RAGBench--Medical \cite{Friel2024RAGBenchEB} is a medical question-answering dataset derived from clinical guidelines and authoritative texts. It contains 1,500 questions spanning fact retrieval and complex reasoning.

\subsection{Facet Decomposition}
\label{sec:facet_decomposition}

We decompose each question into a set of \emph{facets}, as shown in Table~\ref{tab:facet_coverage_bridge}, where each facet represents a minimal semantic unit required for answering the question. We generated the facets as a preprocessing step and held them fixed across all experiments to ensure consistent evaluation as described in Figure \ref{fig:uq_pipeline}.
For a question $q$, facet decomposition produces a set:
\[
\mathcal{F}(q) = \{f_1, f_2, \dots, f_m\},
\]
where each facet $f_i$ is phrased as a standalone question whose answer can be independently validated against retrieved evidence.

For HotpotQA, we leverage the dataset's existing supporting facts (provided as a dataset column)—gold sentences that contain the reasoning chain for each multi-hop question. Following prior work on question decomposition \cite{petcu2025querydecompositionragbalancing, fu-etal-2021-decomposing-complex}, we convert each supporting sentence into interrogative form to create a facet.
For the medical dataset, questions lack annotated supporting facts. We therefore apply GPT-4o-mini with a prompt to decompose each medical question into related facets. This facet decomposition yields 11,007 facets for Medical QA (1,500 questions) and 24,039 facets for HotpotQA (3,000 questions). Each facet is labeled with a semantic type (e.g., temporal, comparative, procedural), which is used in the Results section to analyze evidence usage. See Appendix~\ref{app:complete_example} for a concrete example and Appendix~\ref{app:prompts} for prompt details. The prompt below is used for medical question facet decomposition.

\begin{custompromptbox}{\small Facet Decomposition Prompt}
\setlength{\fboxsep}{3pt}  % Reduce padding (default is 6pt)
\small  % Reduce font size slightly

\textcolor{teal}{\textbf{Task:}} \textit{Given the following medical question, decompose it into 
a list of atomic sub-questions, where each sub-question addresses 
a single clinical fact or reasoning step required to answer the 
original question. Each sub-question should be answerable 
independently using medical knowledge sources.}

\textcolor{blue}{\textbf{Question:}} \texttt{[INPUT\_QUESTION]}

\textcolor{purple}{\textbf{Output format:}} \textit{Numbered list of sub-questions.}
\end{custompromptbox}

\begin{table*}[t]
\centering
\small
\setlength{\tabcolsep}{4pt}
\begin{tabular}{l|ccc|ccc|ccc}
\toprule
Model 
& \multicolumn{3}{c}{Strict RAG}
& \multicolumn{3}{c}{Soft RAG}
& \multicolumn{3}{c}{LLM-only} \\
\cmidrule(lr){2-4} \cmidrule(lr){5-7} \cmidrule(lr){8-10}
& Answer & F1 & BERT
& Answer & F1 & BERT
& Answer & F1 & BERT \\
\midrule

Gemini-2.0-Flash
& Bosnian & 0.00 & 0.00
& Bosnian   & 0.00 & 0.13
& Bosnian   & 0.00 & 0.13 \\

GPT-4o mini
& NO\_ANSWER & 0.00 & 0.00
& Bosnian   & 0.00 & 0.13
& \textbf{Serb} & 1.00 & 1.00 \\

Llama-3-8B-Instruct
& NO\_ANSWER & 0.00 & 0.00
& Bosnian   & 0.00 & 0.13
& The Grand Vizier... & 0.00 & 0.11 \\

\bottomrule
\end{tabular}

\caption{
HotpotQA Example: Generated answers and corresponding F1 and BERTScore for each inference mode.
\textbf{Question:} ``Of what national descent was the Grand Vizier of the Ottoman Empire whose title indicated he had been castrated?''
(\textbf{Gold answer:} \emph{Serb}).
Only GPT in the LLM-only setting produces the correct answer, while Strict and Soft RAG suppress correct parametric knowledge across models.
}
\label{tab:answer_f1_bert_comparison}
\end{table*}

\subsection{Facet$\times$Chunk Evidence Matrix}
\label{sec:facet_chunk}

Given a facet set $\mathcal{F}(q)$ and retrieved chunks $\mathcal{C} = \{c_1, \dots, c_K\}$, we construct a Facet$\times$Chunk matrix that captures facet level coverage over retrieved chunks as shown in Table~\ref{tab:facet_coverage_bridge}.
For each facet–chunk pair $(f_i, c_j)$, we compute the retrieval relevance score $s(f_i, c_j)$ using the retriever’s similarity function. To characterize retrieval uncertainty, we define the retrieval margin:
\[
U_{\text{ret}}(f_i) = s(f_i, c_j^{*}) - s(f_i, c_j^{**}),
\]
where $c_j^{*}$ and $c_j^{**}$ are the top-1 and top-2 retrieved chunks for facet $f_i$. Small margins indicate ambiguous or conflicting retrieval which can cause hallucination in generation side.

To assess whether a generated facet answer $a_i$ is supported by a retrieved chunk $c_j$, we apply a natural language inference (NLI) RoBERTa-large-MNLI model \cite{liu2019roberta} and compute:
\[
\textit{\text{faith}}(a_i, c_j) = P(\text{entail}) - P(\text{contradict}),
\]
yielding a signed grounding score. Positive values indicate evidential support, while negative values indicate contradiction. These retrieval and grounding scores populate the Facet$\times$Chunk matrix and serve as the basis for all subsequent analysis. 

For evidence retrieval, we use a dense retriever based on BGE-base-en-v1.5 embeddings \cite{bge_embedding}, implemented via SentenceTransformers. Source documents are segmented into chunks of 300 tokens with a 50-token overlap, following prior empirical findings that short, fine-grained chunks improve retrieval effectiveness \cite{ramos-varela-etal-2025-context}. For each facet query, we retrieve the top-$K=5$ most similar chunks, balancing recall and precision, as smaller $K$ risks missing relevant evidence while larger $K$ introduces irrelevant or conflicting context \cite{10.5555/3737916.3741766}. Table~\ref{tab:facet_coverage_bridge} shows the Facet$\times$Chunk matrix for a representative question; Appendix~\ref{app:complete_example} provides full matrix construction details, including retrieved chunk content and similarity scores.

\subsection{Controlled Inference Modes}
\label{sec:inference_modes}
To disentangle retrieval failures from generation-level hallucinations, we evaluate three controlled inference modes for each facet:

\textbf{1) Strict RAG}: The model must answer using only retrieved evidence, returning \texttt{NO\_ANSWER} if unsupported:
{\footnotesize
\[
\mathrm{Prompt}_{\text{strict}}(f_i) =
\begin{cases}
\textbf{EVIDENCE: } c_{i1} \oplus c_{i2} \oplus \dots \oplus c_{iK} \\
\textbf{QUESTION: } f_i \\
\textbf{INSTRUCTION: } Evidence Only
\end{cases}
\]
}

\textbf{2) Soft RAG}: The model may mix evidence with prior knowledge:
{\footnotesize
\[
\mathrm{Prompt}_{\text{soft}}(f_i) =
\begin{cases}
\textbf{EVIDENCE: } c_{i1} \oplus \dots \oplus c_{iK} \\
\textbf{QUESTION: } f_i
\end{cases}
\]
}

\textbf{3) LLM-only}: Generation relies solely on parametric model knowledge without retrieval:
{\footnotesize
\[
\mathrm{Prompt}_{\text{lm}}(f_i) =
\begin{cases}
\textbf{QUESTION: } f_i
\end{cases}
\]
}

These modes allow us to distinguish retrieval insufficiency, misleading evidence, and prior-driven reasoning. To illustrate how the three inference modes produce different outputs for the same question, Appendix~\ref{app:complete_example} shows a complete example. This demonstrates how comparing the three modes enables causal attribution of hallucination sources. All LLM prompts are provided in Appendix~\ref{app:prompts}.

\begin{table*}[t]
\centering
\small
\setlength{\tabcolsep}{9.5pt}
\begin{tabular}{lrrrrrr}
\toprule
\multirow{2}{*}{\textbf{Models}} & \multicolumn{5}{c}{\textbf{Evidence Taxonomy}} \\
 & Failure & Overridden & Misalignment & Helpful & Robust \\
\midrule
Gemini-2.0 Flash & 20.6 & 12.5 & 22.3 & 9.6 & 35.1 \\
GPT-4o mini & 0.2 & 34.4 & 0.2 & 51.6 & 13.7 \\
Llama-3-8B-Instruct & 0.1 & 38.2 & 0.2 & 53.1 & 8.3  \\
\midrule
Overall & 7.0 & 28.4 & 7.5 & 38.1 & 19.0 \\
\bottomrule
\end{tabular}
\caption{Medical Dataset: Evidence Taxonomy Distribution Across Models. Percentage of medical reasoning facets (N=11,007) in each evidence category. \textbf{Overall Evidence Helpful (38.1\%) demonstrates that domain-specific medical retrieval frequently corrects model errors.} Gemini exhibits balanced behavior across failure and override, while GPT and LLaMA demonstrate override-dominant patterns.}
\label{tab:medical_taxonomy}
\end{table*}

\begin{table*}[t]
\centering
\small
\begin{tabular}{llcccccc}
\toprule
\multirow{2}{*}{Question Type} & \multirow{2}{*}{Model} &
\multicolumn{2}{c}{Strict RAG} &
\multicolumn{2}{c}{Soft RAG} &
\multicolumn{2}{c}{LLM-only} \\
\cmidrule(lr){3-4} \cmidrule(lr){5-6} \cmidrule(lr){7-8}
 & & F1 & BERT & F1 & BERT & F1 & BERT \\
\midrule
\multirow{3}{*}{Fact Retrieval}
 & Gemini-2.0-Flash & 0.222 & 0.093 & \textbf{0.376} & \textbf{0.375} & \textbf{0.337} & 0.351 \\
 & GPT-4o mini    & 0.185 & 0.008 & 0.232 & 0.139 & 0.186 & 0.101 \\
 & Llama-3-8B-Instruct  & \textbf{0.291} & \textbf{0.231} & 0.351 & 0.317 & 0.309 & 0.251 \\
\midrule
\multirow{3}{*}{Complex Reasoning}
 & Gemini-2.0-flash & 0.176 & 0.034 & 0.357 & 0.360 & 0.340 & \textbf{0.337} \\
 & GPT-4o mini    & 0.089 & 0.025 & 0.161 & 0.116 & 0.127 & 0.087 \\
 & Llama-3-8B-Instruct  & 0.226 & 0.175 & 0.306 & 0.298 & 0.306 & 0.296 \\
\bottomrule
\end{tabular}
\caption{Medical Dataset: Mean F1 and BERTScore by question type and inference mode. In medical dataset, Soft RAG outperforms Strict RAG across models, especially for Complex Reasoning}
\label{tab:medical_qa_type}
\end{table*}

\subsection{Facet-Level Evidence Taxonomy}
\label{sec:taxonomy}
To understand how retrieval and generation contribute to hallucination, we develop a \emph{Facet-Level Evidence Taxonomy} that classifies each reasoning facet based on evidence availability and usage patterns. Each facet receives one taxonomy label.

For a facet $f_i$ with retrieved evidence chunks $C_i = \{c_{i1}, \dots, c_{iK}\}$ and generated answer $a_i$, we assess whether evidence is retrieved ($R(c_{ij}, f_i) \in \{0,1\}$) and whether the generated answer is grounded in that evidence. For each answer-chunk pair, we compute an NLI grounding score using RoBERTa-large-MNLI \cite{liu2019roberta}, where positive values indicate supportive evidence and negative values indicate contradiction.

Each facet is classified by comparing the success/failure patterns across the three controlled inference modes described in Section~\ref{sec:inference_modes}. Using this framework, we define the following evidence taxonomy categories:

\begin{itemize}
\setlength{\itemsep}{0pt}
\setlength{\parskip}{0pt}
\setlength{\parsep}{0pt}
\setlength{\topsep}{1pt}
    \item \textbf{Evidence Failure:} Retrieval finds no relevant information for the facet. Strict RAG returns \texttt{NO\_ANSWER}.
    \item \textbf{Evidence Misalignment:} Retrieved chunks are topically related but contradict or fail to support the generated facet answer.
    \item \textbf{Evidence Overridden:} Relevant evidence is retrieved, but the LLM ignores it and relies on parametric knowledge instead. Strict RAG fails, while Soft RAG or LLM-only succeeds, indicating evidence override during answer generation.
    \item \textbf{Evidence Helpful:} Retrieved evidence corrects LLM errors. LLM-only fails, but Strict RAG succeeds, demonstrating that retrieval reduces hallucination.
    \item \textbf{Evidence Robust:} All three inference modes produce correct and consistent answers. The facet answer remains stable regardless of evidence availability.
\end{itemize}

All taxonomy assignments are performed independently at the facet level, enabling aggregation of facet-level evidence patterns to analyze question-level behavior in subsequent sections. Appendix~\ref{app:complete_example} provides a detailed example of the \textit{Evidence Overridden} pattern.

\subsection{Question-Level Aggregation}
\label{sec:question_level}

While our primary analysis operates at the facet level, users interact with final answers. We therefore aggregate facet-level outputs into a single question-level answer under each inference mode.

Question-level aggregation proceeds as follows: For each inference mode, we collect the generated facet answers $\{a_1, a_2, \ldots, a_m\}$ corresponding to facets $\{f_1, f_2, \ldots, f_m\}$. We then prompt the same LLM to synthesize these facet answers into a final question-level answer. The complete aggregation prompt is provided in Appendix~\ref{app:prompts}.

The final answer is evaluated using token-level F1 score \cite{rajpurkar2016squad} and BERTScore \cite{zhang2020bertscore}, which are standard metrics for evaluating question answering \cite{xlqa2025}. F1 measures the average token overlap between predicted and ground truth answers, balancing precision and recall. BERTScore captures semantic similarity through contextualized embeddings, addressing limitations of pure lexical overlap. 

Table~\ref{tab:answer_f1_bert_comparison} shows question-level answer generation across all three inference modes for a representative example. Specifically, it illustrates a case where Strict RAG returns \texttt{NO\_ANSWER}, Soft RAG produces an incorrect answer that contradicts the retrieved evidence, and LLM-only successfully generates the correct answer using parametric knowledge alone. These metrics contextualize how facet-level evidence behaviors translate into end-to-end answer quality rather than defining hallucination. All evaluations are conducted consistently across Gemini, GPT, and LLaMA models under identical inference settings listed in Appendix~\ref{app:configuration}.

\section{Experimental Results}
\label{main:results}

We analyze hallucination behavior in RAG across three models (Gemini, GPT, LLaMA) on two datasets Medical QA and HotpotQA. Our analysis reveals a weak and inconsistent relationship between retrieval quality and generation faithfulness: high-quality retrieval frequently co-occurs with unfaithful generation, while faithful answers arise even under imperfect retrieval. This decoupling motivates facet-level analysis of how retrieved evidence is actually used during generation.

\subsection{Evidence Taxonomy Results}

Table~\ref{tab:medical_taxonomy} presents the distribution of medical reasoning facets across five evidence categories, revealing structured failure patterns rather than random hallucination. \textbf{Evidence Override emerges as the dominant failure mode at 28.4\%}, where models systematically ignore retrieved evidence in favor of parametric priors. This is 4× more prevalent than Evidence Failure (7.0\%), \textbf{indicating that the primary challenge is not finding evidence but ensuring models use it during generation}. Evidence Helpful (38.1\%) demonstrates that domain-specific retrieval frequently corrects model errors, validating RAG's core benefit. Combined with Evidence Robust (19.0\%), successful evidence utilization accounts for 57.1\% of facets.

LLM-specific patterns diverge substantially. Gemini exhibits balanced behavior with 20.6\% failure and 12.5\% override, alongside highest robustness (35.1\%). In contrast, GPT and LLaMA demonstrate override-dominant patterns (34.4\% and 38.2\% respectively) with near-zero failure rates (0.2\% and 0.1\%), indicating strong retrieval but systematic parametric interference that undermines evidence usage. To validate cross-domain generalization, we analyze HotpotQA (Table~\ref{tab:hotpot_taxonomy}). \textbf{Evidence Override remains dominant at 42.3\%}, now exceeding medical domain's 28.4\%. This increase occurs because HotpotQA's general-domain content has high coverage in LLM pretraining corpora, strengthening parametric priors that override retrieved evidence. Evidence Failure drops to 5.8\% (versus 7.0\% for medical), while Evidence Misalignment nearly vanishes (0.7\% versus 7.5\% medical), confirming Wikipedia's broad coverage provides better retrieval recall. The consistent 7:1 override-to-failure ratio across datasets (42.3\% vs 5.8\% HotpotQA, 28.4\% vs 7.0\% medical) demonstrates that retrieval-generation misalignment is a fundamental architectural limitation rather than domain-specific artifact.

\begin{figure}  % Capital H forces HERE
    \centering
    \includegraphics[width=0.45\textwidth,height=0.18\textheight]{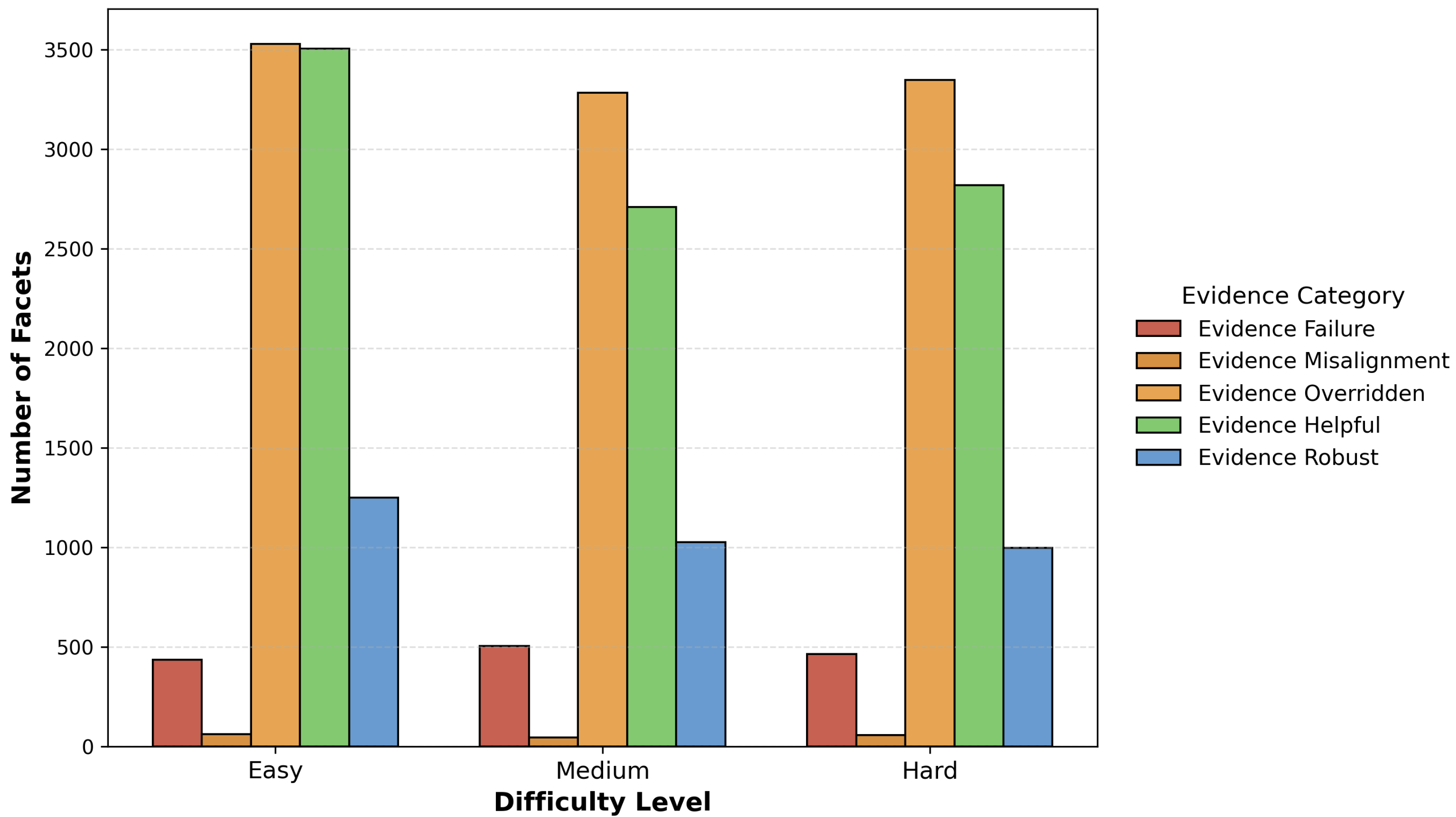}
    \caption{HotpotQA: Evidence Taxonomy Distribution by Difficulty Level. Override and Helpful are balanced on easy questions, but Helpful declines on harder questions while Override stays stable. Failure and Misalignment remain constant and minimal, while Evidence Robust decreases with difficulty, \textbf{indicating complexity disrupts generation rather than retrieval}.}
    \label{fig:hotpot_difficulty_taxonomy}
    \end{figure}    

\begin{table*}[t]
\centering
\small
\setlength{\tabcolsep}{9.5pt}
\begin{tabular}{lrrrrrr}
\toprule
\multirow{2}{*}{\textbf{Models}} & \multicolumn{5}{c}{\textbf{Evidence Taxonomy}} \\
 & Failure & Overridden & Misalignment & Helpful & Robust \\
% Models & Evidence Failure & Evidence Overridden & Evidence Misalign. & Evidence Helpful & Evidence Robust \\
\midrule
Gemini-2.0 Flash & 14.7 & 32.1 & 0.7 & 34.5 & 18.0\\
GPT-4o mini & 2.8 & 43.6 & 0.7 & 37.3 & 15.6\\
Llama-3-8B-Instruct & 2.0 & 47.1 & 2.7 & 41.0 & 7.2\\
\midrule
Overall & 5.8 & 42.3 & 0.7 & 37.6 & 13.6\\
\bottomrule
\end{tabular}
\caption{HotpotQA: Evidence Taxonomy Distribution Across Models. Distribution of reasoning facets (N=24,039) by evidence behavior. \textbf{Evidence Override (42.3\%) exceeds Failure (5.8\%) by 7×, confirming generation-side grounding challenges persist across domains.}}
\label{tab:hotpot_taxonomy}
\end{table*}

\begin{table*}[t]
\centering
\small
\setlength{\tabcolsep}{5pt}
\begin{tabular}{llrrrrrr}
\toprule
\textbf{Question Difficulty} & \textbf{Model} &
\multicolumn{2}{c}{\textbf{Strict RAG}} &
\multicolumn{2}{c}{\textbf{Soft RAG}} &
\multicolumn{2}{c}{\textbf{LLM-only}} \\
\cmidrule(lr){3-4} \cmidrule(lr){5-6} \cmidrule(lr){7-8}
 &  & F1 & BERT & F1 & BERT & F1 & BERT \\
\midrule
\multirow{3}{*}{Easy}
 & Gemini-2.0-Flash
 & \textbf{0.550} & 0.443 & 0.624 & 0.539 & 0.564 & 0.544 \\
 & GPT-4o-mini
 & 0.519 & \textbf{0.447} & 0.613 & 0.558 & 0.498 & 0.477 \\
 & LLaMA-3-8B-Instruct
 & 0.196 & 0.026 & 0.315 & 0.174 & 0.230 & 0.108 \\
\midrule
\multirow{3}{*}{Medium}
 & Gemini-2.0-Flash
 & 0.528 & 0.429 & 0.687 & 0.624 & \textbf{0.582} & \textbf{0.623} \\
 & GPT-4o-mini
 & 0.501 & 0.389 & \textbf{0.699} & \textbf{0.631} & 0.504 & 0.518 \\
 & LLaMA-3-8B-Instruct
 & 0.196 & 0.033 & 0.351 & 0.236 & 0.182 & 0.053 \\
\midrule
\multirow{3}{*}{Hard}
 & Gemini-2.0-Flash
 & 0.473 & 0.349 & 0.579 & 0.507 & 0.515 & 0.516 \\
 & GPT-4o-mini
 & 0.411 & 0.300 & 0.609 & 0.549 & 0.427 & 0.431 \\
 & LLaMA-3-8B-Instruct
 & 0.149 & 0.131 & 0.296 & 0.171 & 0.195 & 0.059 \\
\bottomrule
\end{tabular}
\caption{HotpotQA: Question-Level by Difficulty. Mean F1 and BERTScore by difficulty and inference mode. Soft RAG outperforms Strict RAG, and performance peaks at medium difficulty.}
\label{tab:hotpot_difficulty}
\end{table*}

\begin{table*}
\centering
\small
\begin{tabular}{llrrrrrr}
\toprule
\textbf{Question Type} & \textbf{Model} &
\multicolumn{2}{c}{\textbf{Strict RAG}} &
\multicolumn{2}{c}{\textbf{Soft RAG}} &
\multicolumn{2}{c}{\textbf{LLM-only}} \\
\cmidrule(lr){3-4} \cmidrule(lr){5-6} \cmidrule(lr){7-8}
 &  & F1 & BERT & F1 & BERT & F1 & BERT \\
\midrule
\multirow{3}{*}{Bridge}
 & Gemini-2.0-Flash
 & \textbf{0.561} & \textbf{0.447} & 0.631 & 0.553 & 0.476 & 0.483 \\
 & GPT-4o mini
 & 0.483 & 0.350 & \textbf{0.655} & \textbf{0.605} & 0.375 & 0.374 \\
 & LLaMA-3-8B-Instruct
 & 0.222 & 0.135 & 0.369 & 0.240 & 0.170 & 0.416 \\
\midrule
\multirow{3}{*}{Comparison}
 & Gemini-2.0-Flash
 & 0.472 & 0.377 & 0.614 & 0.547 & \textbf{0.609} & \textbf{0.614} \\
 & GPT-4o mini
 & 0.470 & 0.417 & 0.603 & 0.544 & 0.561 & 0.563 \\
 & LLaMA-3-8B-Instruct
 & 0.119 & 0.137 & 0.234 & 0.099 & 0.231 & 0.105 \\
\bottomrule
\end{tabular}
\caption{HotpotQA: Question-level answer quality by question meta-type. Mean F1 and BERTScore are reported across inference modes.}
\label{tab:hotpot_question_type}
\end{table*}

\begin{figure}
    \centering
    \includegraphics[width=0.45\textwidth,height=0.15\textheight]{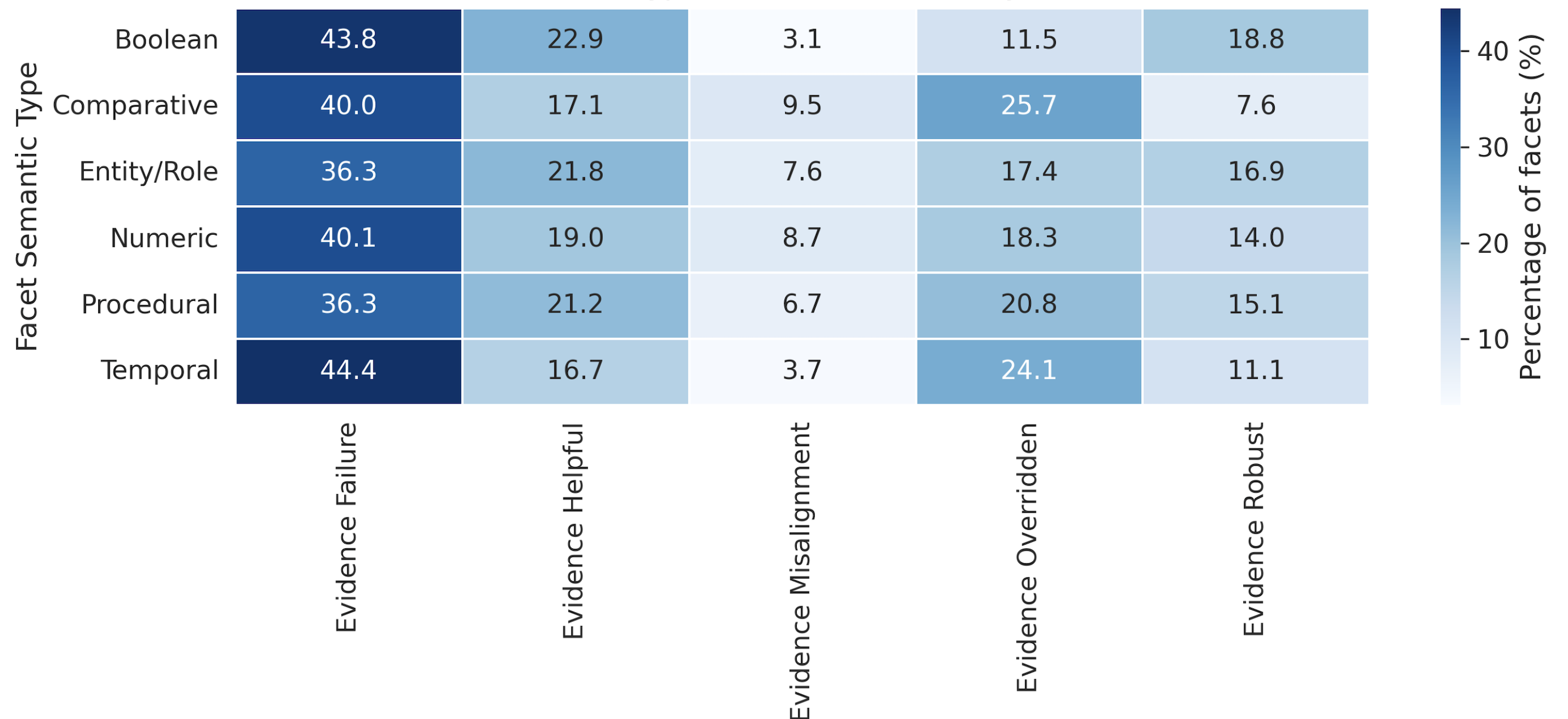}
   \caption{Medical Dataset: Facet semantic type × evidence taxonomy distribution. Boolean and Temporal facets show highest failure rates. Comparative facets are most unstable with highest misalignment and lowest robust rates.}
    \label{fig:medical_semantic}
\end{figure}
\subsection{Facet-Level Evidence Patterns}

Figure~\ref{fig:hotpot_difficulty_taxonomy} shows key evidence patterns across difficulty levels in HotpotQA. Easy questions show a 50--50 balance between Evidence Override and Evidence Helpful, providing benefit half the time. As difficulty increases, this balance breaks---but not as expected. \textbf{While we expected harder questions to cause more retrieval failures, Evidence Failure remains constant at $\sim$6\% and Evidence Misalignment stays negligible at $<$2\% across all levels}. Instead, Override stays stable at $\sim$40--43\% while Helpful drops to $\sim$35\%, showing increased complexity reduces the LLM’s ability to integrate available evidence. Evidence Robust decreases from 14\% (easy) to 13\% (hard), as increased complexity disrupts consistent reasoning even when parametric knowledge and evidence align. \textbf{The stable failure rate indicates that retrieval finds relevant evidence}, yet the declining Helpful category shows that models increasingly fail to use that evidence correctly during generation.

\begin{figure}
    \centering
    \includegraphics[width=0.48\textwidth,height=0.15\textheight]{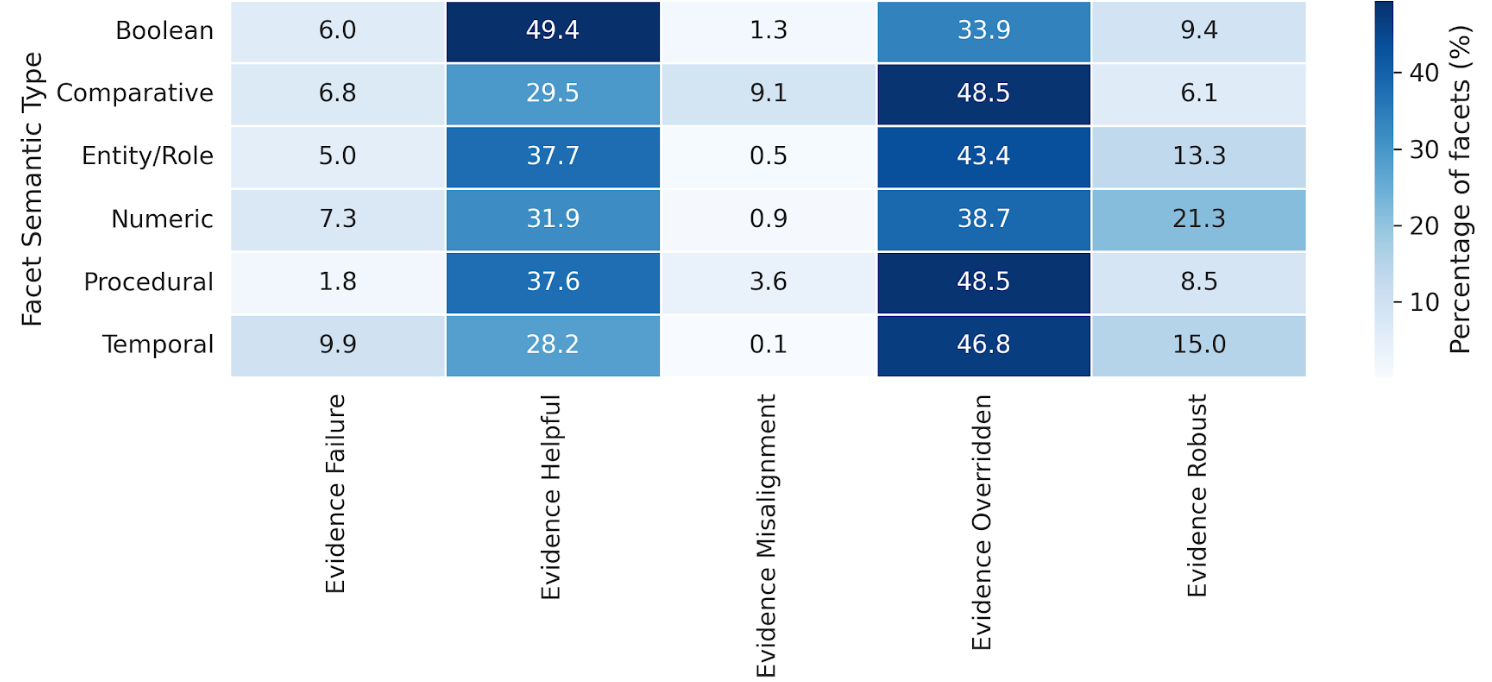}
    \caption{HotpotQA: Facet Semantic Type × Evidence Taxonomy Distribution. Boolean and Temporal facets show lowest failure rates. Override rates are consistently high across all types. Comparative facets remain most unstable in both datasets.}
    \label{fig:hotpot_semantic}
\end{figure}

Figures ~\ref{fig:medical_semantic} and ~\ref{fig:hotpot_semantic} examine evidence patterns across different facet semantic types (comparison, numerical, procedural, temporal, ...), revealing domain-specific retrieval but consistent generation behavior. Failure rates vary dramatically: Boolean and Temporal facets fail most in medical QA (43.8\%, 44.4\%) but least in HotpotQA (6.0\%, 9.9\%). Override rates nearly double from medical to HotpotQA, indicating stronger parametric priors for general knowledge. However, Comparative facets remain most unstable in both datasets (highest misalignment ~9\%, lowest robust ~7\%). Critically, misalignment stays below 10\% across all types in both datasets, confirming generation-side override dominates over retrieval-side contradictions. Faithfulness Analysis: Evidence Helpful and Robust achieve high faithfulness (median $>$ 0.7), while Override shows bimodal behavior spanning from high support to contradiction. This explains why retrieval confidence cannot predict generation quality: models unpredictably either incorporate or contradict retrieved evidence. Detailed distributions in Appendix~\ref{app:results} Figure~\ref{fig:medical_faithfulness}.

\subsection{Question-Level Performance}

Table~\ref{tab:medical_qa_type} presents mean F1 and BERTScore across inference modes on medical QA. Soft RAG substantially outperforms Strict RAG, with gains varying by model and question type. Complex reasoning benefits most: Gemini +103\% (0.176 → 0.357 F1), GPT +81\% (0.089 → 0.161 F1), and LLaMA +35\% (0.226 → 0.306 F1). Fact retrieval shows more moderate gains across models.

Table~\ref{tab:hotpot_difficulty} shows question-level results on HotpotQA stratified by difficulty. Soft RAG consistently outperforms Strict RAG across all difficulty levels: easy questions show 13--61\% gains, medium questions show 30--79\% gains, and hard questions show 22--99\% gains. \textbf{Medium-difficulty questions achieve the highest performance across models, indicating that question decomposition and evidence integration are most effective at this level}.

Table~\ref{tab:hotpot_question_type} presents results by HotpotQA question meta-type. Soft RAG outperforms Strict RAG across both Bridge and Comparison questions. Bridge questions, which require connecting information across documents, show particularly strong gains for GPT (0.483 → 0.655 F1, +36\%). Comparison questions exhibit more balanced performance, with LLM-only mode approaching Soft RAG performance (Gemini: 0.609 vs.\ 0.614 F1), suggesting that comparative reasoning may be well-captured in parametric knowledge.

\textbf{Findings}: LLM-only mode frequently exceeds Strict RAG in both datasets (Medical: Gemini 0.340 vs.\ 0.176 F1 for complex reasoning; HotpotQA: Gemini 0.515 vs.\ 0.473 F1 for hard questions), underscoring that performance differences are driven by evidence utilization rather than knowledge availability. However, aggregate metrics mask substantial per-question variability: Appendix~\ref{app:results} shows that approximately 30\% of questions degrade under Soft RAG, highlighting the lack of robust mechanisms for deciding when to rely on retrieved evidence versus parametric knowledge. Model rankings remain consistent, with Gemini achieving the strongest Soft RAG performance, followed by GPT and LLaMA.

\subsection{Summary}

Our results demonstrate that hallucination in RAG arises from systematic retrieval-generation misalignment rather than retrieval failure alone. Evidence Override dominates (28.4\% medical, 42.3\% HotpotQA), exceeding Evidence Failure by 4-7× (7.0\% medical, 5.8\% HotpotQA). \textbf{This ratio indicates the primary reason lies in generation-side grounding: relevant evidence is often successfully retrieved but ignored during generation}. Question-level performance confirms these patterns: Soft RAG improves F1 by allowing evidence flexibility, while strict grounding degrades performance below no-retrieval baseline in 30\% of cases. Cross-dataset validation reveals that override patterns persist regardless of knowledge domain (specialized medical vs.\ open Wikipedia), question structure, and difficulty level. These findings identify retrieval–generation misalignment as a fundamental architectural limitation and demonstrate that \textit{improving RAG requires not only better retrieval but also mechanisms to ensure retrieved evidence is integrated correctly and adaptively}.

\section{Conclusion}
Our key finding overturns a foundational assumption: better retrieval does not guarantee better answers generation. Across medical QA and HotpotQA datasets, models consistently override retrieved evidence far more often than retrieval fails. This pattern holds across three LLM families (GPT, Gemini, and LLaMA), multiple difficulty levels, and distinct knowledge domains. Strict evidence grounding often underperforms generation without retrieval entirely, revealing that current RAG architectures lack mechanisms to balance external evidence against parametric knowledge. By decomposing reasoning into atomic facets and tracing evidence usage at each step, our facet-level diagnostic framework provides an interpretable view of retrieval–generation behavior. This positions facet-level analysis as a first step toward explainable and uncertainty-aware RAG systems, enabling future models to make evidence integration decisions more transparent, adaptive, and controllable.

\section{Limitations}
This work is diagnostic in nature and does not propose new retrieval or generation models. While our framework enables fine-grained analysis of retrieval-generation misalignment, it relies on facet decomposition and NLI model quality, which may introduce noise or bias. Our experiments are limited to two datasets (medical QA and HotpotQA) and three LLMs; observed patterns may differ for other tasks, languages, or architectures. Additionally, our analysis is limited to offline evaluation and does not study interactive or user-facing RAG systems. We view this as a first step toward explainable RAG systems: by making evidence usage explicit at the facet level, our framework provides transparent diagnostic signals. Future work can build on this foundation to develop interactive, uncertainty-aware RAG systems that adaptively balance external evidence with parametric knowledge.

\section{Acknowledgments}
We used generative AI tools to assist with language polishing and clarity of presentation. All scientific content, experimental design, analysis, and conclusions were developed and verified by the authors.

\section{Ethical considerations}
\paragraph{Datasets and Data Usage.} 
This work uses publicly available datasets: HotpotQA \cite{yang2018hotpotqa} and RAGBench–Medical \cite{Friel2024RAGBenchEB}. The medical dataset is derived from published clinical guidelines and authoritative medical texts, not patient data. No personally identifiable information or protected health information was used. All datasets were accessed through their official repositories and used in accordance with their licenses.

\paragraph{Model Access and Reproducibility.} 
Our evaluation uses three LLMs: GPT-4o-mini (via OpenAI API), Gemini-2.0-Flash (via Google Generative AI API), and LLaMA-3-8B-Instruct (open-weight model). Access to commercial APIs requires appropriate subscriptions. We provide detailed model configurations \ref{app:configuration} and prompt templates \ref{app:prompts} to support reproducibility. Our diagnostic framework does not train new models and therefore does not introduce additional environmental costs beyond inference.

\paragraph{Broader Impacts.} 
By exposing systematic retrieval-generation misalignment, this work may benefit developers of RAG systems in reducing hallucination risks. However, improved hallucination detection does not eliminate all risks of LLM deployment. Our findings that models frequently override retrieved evidence have implications for trust calibration in human-AI systems: users may over-rely on RAG systems assuming retrieval guarantees faithfulness. We encourage practitioners to implement appropriate uncertainty quantification and human-in-the-loop verification, especially in high-stakes applications.

\bibliography{custom}

\appendix

\section{Illustrative Facet-Level Diagnostic Example}
\label{app:complete_example}
This section illustrates our facet-level diagnostic framework using a HotpotQA example. Despite retrieving relevant evidence with high similarity scores, the RAG system could not faithfully connect the Oklahoma player in the 1951 Sugar Bowl with the Heisman Trophy winner. 

This demonstrates the "Evidence Overridden" category: Strict RAG failed completely, Soft RAG bridged the gap using parametric knowledge, and LLM-only succeeded with prior knowledge alone—revealing that retrieval quality alone does not guarantee faithful evidence utilization.

The question (HotpotQA): ``Which Oklahoma player in the 1951 Sugar Bowl went on to win the Heisman Trophy?'
  
\begin{figure}[H]
  \centering
  \scriptsize  % Even smaller font
  \begin{tcolorbox}[colback=gray!5!white,colframe=black!75!black,boxsep=3pt,left=3pt,right=3pt,top=8pt,bottom=8pt]
  \textbf{Example: Facet-RAG Complete Pipeline}
  
  \textbf{Original Question (HotpotQA):} ``Which Oklahoma player in the 1951 Sugar Bowl went on to win the Heisman Trophy?'' \textbf{Gold Answer:} Billy Vessels
  
  \textbf{Step 1: Facet Decomposition}
  \begin{itemize}
  \setlength{\itemsep}{0pt}
\setlength{\parskip}{0pt}
\setlength{\parsep}{0pt}
\setlength{\topsep}{2pt}
  \item Facet 1: ``Which player from Oklahoma participated in the 1951 Sugar Bowl?''
  \item Facet 2: ``Did Billy Vessels win the Heisman Trophy?''
  \end{itemize}
  
  \textbf{Step 2: Chunk Retrieval} (Top-5 per facet, showing top-3 as sample)

  \begin{itemize}
  \setlength{\itemsep}{0pt}
\setlength{\parskip}{0pt}
\setlength{\parsep}{0pt}
\setlength{\topsep}{2pt}
  \item Facet 1 most relevant: C2(0.82), C1(0.72), C3(0.72) 
  \item Facet 2 most relevant: C8(0.77), C9(0.77), C10(0.74)
  \end{itemize}
  
  \textbf{Step 3: NLI-Based Faithfulness Calculation}
  
  For each facet answer (hypothesis), run NLI vs all chunks (premises): $\text{faithfulness} = P(\text{entailment}) - P(\text{contradiction})$. Facet faithfulness = max over all chunks.
  
  \textbf{Step 4: Facet-Level Answer Generation using prompt in Appendix \ref{app:prompts}:}
  
  \textit{Facet 1: ``Which player from Oklahoma participated in the 1951 Sugar Bowl?''}
  \begin{itemize}
  \setlength{\itemsep}{0pt}
\setlength{\parskip}{0pt}
\setlength{\parsep}{0pt}
\setlength{\topsep}{2pt}
  \item \textbf{Strict RAG:} ``NO\_ANSWER''
  \item \textbf{Soft RAG:} ``The player from Oklahoma who participated in the 1951 Sugar Bowl was Heisman Trophy winner Billy Vessels.''
  \item \textbf{LLM-only:} ``The player from Oklahoma who participated in the 1951 Sugar Bowl was Billy Vessels.''
  \end{itemize}
  
  \textit{Facet 2: ``Did Billy Vessels win the Heisman Trophy?''}
  \begin{itemize}
  \setlength{\itemsep}{0pt}
\setlength{\parskip}{0pt}
\setlength{\parsep}{0pt}
\setlength{\topsep}{2pt}
  \item \textbf{Strict RAG:} ``NO\_ANSWER''
  \item \textbf{Soft RAG:} ``Yes, Billy Vessels won the Heisman Trophy in 1952.''
  \item \textbf{LLM-only:} ``Yes, Billy Vessels won the Heisman Trophy in 1952.''
  \end{itemize}
  
  \textbf{Step 5: NLI-Based Faithfulness Calculation}
  
  Facet 1 faithfulness\_soft = 0.00 (max over chunks), and Facet 2 faithfulness\_soft = 0.103 (max from C10)
  
  \textbf{Step 6: Facet-Level Evidence Taxonomy Assignment}
  
  \textbf{Facet 1:} \textbf{Evidence Overridden} -- Reason: Strong retrieval (max sim: 0.82) but faithfulness = 0.00. Strict RAG fails; Soft RAG relies on parametric knowledge.
  
  \textbf{Facet 2:} \textbf{Evidence Overridden} -- Reason: Strong retrieval (max sim: 0.77) but faithfulness = 0.103. Retrieved evidence insufficient to fully support facet answer.
  
  \textbf{Step 7: Question-Level Answer Aggregation}
  
  Facet answers aggregated to full question answers using prompt in Appendix \ref{app:prompts}:
  \begin{itemize}
  \setlength{\itemsep}{0pt}
\setlength{\parskip}{0pt}
\setlength{\parsep}{0pt}
\setlength{\topsep}{2pt}
  \item \textbf{Strict RAG:} ``NO\_ANSWER'' (F1 = 0.00) -- Both facets returned NO\_ANSWER
  \item \textbf{Soft RAG:} ``The Oklahoma player who participated in the 1951 Sugar Bowl and went on to win the Heisman Trophy is Billy Vessels. He won the award in 1952.'' (F1 = 0.16) -- Partial match
  \item \textbf{LLM-only:} ``Billy Vessels.'' (F1 = 1.00) -- Exact match
  \item \textbf{Question Category:} Evidence Overridden (both facets classified as Evidence Overridden)
  \end{itemize}
  \end{tcolorbox}
  \caption{Illustrative facet-level diagnostic example from HotpotQA.}
  \label{fig:facet_level_example_sugar_bowl}
\end{figure}

\section{Prompts and Implementation Details}
\label{app:prompts}
\subsection{Facet Decomposition Prompt (Medical Dataset)}

For medical questions without supporting fact annotations:

\begin{custompromptbox}{Facet Decomposition Prompt}
\setlength{\fboxsep}{3pt}  % Reduce padding (default is 6pt)
\small  % Reduce font size slightly

\textcolor{teal}{\textbf{Task:}} \textit{Given the following medical question, decompose it into 
a list of atomic sub-questions, where each sub-question addresses 
a single clinical fact or reasoning step required to answer the 
original question. Each sub-question should be answerable 
independently using medical knowledge sources.}

\textcolor{blue}{\textbf{Question:}} \texttt{[INPUT\_QUESTION]}

\textcolor{purple}{\textbf{Output format:}} \textit{Numbered list of sub-questions.}
\end{custompromptbox}

\subsection{Facet Decomposition Prompt (HotpotQA Dataset)}

For HotpotQA, we leverage the dataset's existing supporting facts annotations. We use few-shot prompting to convert supporting sentences into interrogative facets. We design question-type-specific prompts:

\textbf{1) Bridge Questions Prompt:}

\begin{custompromptbox}{Facet Decomposition Prompt}
\setlength{\fboxsep}{3pt}
\small
\textcolor{teal}{\textbf{Task:}} \textit{Convert this bridge question into reasoning steps (facets).}

\textbf{Example 1:}\\
\textcolor{blue}{\textbf{Question:}} What nationality is the director of the film Masked and Anonymous?\\
Supporting Facts: \texttt{[["Masked and Anonymous", 0], ["Larry Charles", 0]]}\\
\textcolor{purple}{\textbf{Facets:}}
\begin{enumerate}
    \item Who directed the film Masked and Anonymous?
    \item What is Larry Charles's nationality?
\end{enumerate}

\textbf{Example 2:}\\
\textcolor{blue}{\textbf{Question:}} What year was the director of Blade Runner born?\\
Supporting Facts: \texttt{[["Blade Runner", 1], ["Ridley Scott", 0]]}\\
\textcolor{purple}{\textbf{Facets:}}
\begin{enumerate}
    \item Who directed Blade Runner?
    \item When was Ridley Scott born?
\end{enumerate}

\textbf{Now convert this:}\\
\textcolor{blue}{\textbf{Question:}} \texttt{[INPUT\_QUESTION]}\\
Supporting Facts: \texttt{[INPUT\_FACTS]}\\
\textcolor{purple}{\textbf{Facets:}}
\end{custompromptbox}

\textbf{2) Comparison Questions Prompt:}

\begin{custompromptbox}{Facet Decomposition Prompt}
\setlength{\fboxsep}{3pt}
\small
\textcolor{teal}{\textbf{Task:}} \textit{Convert this comparison question into reasoning steps (facets).}

\textbf{Example 1:}\\
\textcolor{blue}{\textbf{Question:}} Who was born first, Arthur Conan Doyle or Artur Schnitzler?\\
Supporting Facts: \texttt{[["Arthur Conan Doyle", 0], ["Artur Schnitzler", 0]]}\\
\textcolor{purple}{\textbf{Facets:}}
\begin{enumerate}
    \item When was Arthur Conan Doyle born?
    \item When was Artur Schnitzler born?
\end{enumerate}

\textbf{Example 2:}\\
\textcolor{blue}{\textbf{Question:}} Which has more species, genus A or genus B?\\
Supporting Facts: \texttt{[["Genus A", 0], ["Genus B", 0]]}\\
\textcolor{purple}{\textbf{Facets:}}
\begin{enumerate}
    \item How many species are in genus A?
    \item How many species are in genus B?
\end{enumerate}

\textbf{Now convert this:}\\
\textcolor{blue}{\textbf{Question:}} \texttt{[INPUT\_QUESTION]}\\
Supporting Facts: \texttt{[INPUT\_FACTS]}\\
\textcolor{purple}{\textbf{Facets:}}
\end{custompromptbox}

\subsection{Facet-Level Answer Generation Prompts}

We generate answers for each reasoning facet under three controlled inference modes:

\subsubsection{Strict RAG Prompt}

For facet-level generation with strict evidence grounding:

\begin{custompromptbox}{Strict RAG Facet Generation}
\setlength{\fboxsep}{3pt}  % Reduce padding (default is 6pt)
\small  % Reduce font size slightly

\textcolor{teal}{\textbf{System:}} \textit{You are a factual question answering assistant. Give a short, direct answer in one or two sentences. State facts clearly without background explanation.}

\textcolor{red}{\textbf{Constraint:}} \textit{Use ONLY the evidence below to answer the question. If the evidence does not clearly support an answer, respond exactly with} \texttt{NO\_ANSWER}\textit{.}

\textcolor{teal}{\textbf{Evidence:}} 
\begin{itemize}
\item \texttt{[CHUNK\_1]}
\item \texttt{[CHUNK\_2]}
\item ...
\item \texttt{[CHUNK\_K]}
\end{itemize}

\textcolor{blue}{\textbf{Question:}} \texttt{[FACET]}

\textcolor{purple}{\textbf{Output:}} \textit{Short answer (1-2 sentences) or} \texttt{NO\_ANSWER}
\end{custompromptbox}

\subsubsection{Soft RAG Prompt}

For flexible facet-level generation allowing parametric knowledge integration:

\begin{custompromptbox}{Soft RAG Facet Generation}
\setlength{\fboxsep}{3pt}  % Reduce padding (default is 6pt)
\small  % Reduce font size slightly

\textcolor{teal}{\textbf{System:}} \textit{You are a factual question answering assistant. Give a short, direct answer in one or two sentences. State facts clearly without background explanation.}

\textcolor{teal}{\textbf{Instruction:}} \textit{You may use the evidence below and your general knowledge.}

\textcolor{teal}{\textbf{Evidence:}}
\begin{itemize}
\item \texttt{[CHUNK\_1]}
\item \texttt{[CHUNK\_2]}
\item ...
\item \texttt{[CHUNK\_K]}
\end{itemize}

\textcolor{blue}{\textbf{Question:}} \texttt{[FACET]}

\textcolor{purple}{\textbf{Output:}} \textit{Short answer (1-2 sentences)}
\end{custompromptbox}

\subsubsection{LLM-only Prompt}

For facet-level generation without retrieval:

\begin{custompromptbox}{LLM-only Facet Generation}
\setlength{\fboxsep}{3pt}  % Reduce padding (default is 6pt)
\small  % Reduce font size slightly

\textcolor{teal}{\textbf{System:}} \textit{You are a factual question answering assistant. Give a short, direct answer in one or two sentences. State facts clearly without background explanation.}

\textcolor{blue}{\textbf{Question:}} \texttt{[FACET]}

\textcolor{purple}{\textbf{Output:}} \textit{Short answer (1-2 sentences)}
\end{custompromptbox}

No external evidence is provided. The model relies entirely on parametric knowledge.

\subsection{Question-Level Aggregation Prompt}

To aggregate facet-level answers into a final question-level answer, we use the following prompt for Soft RAG; the remaining modes follow the same procedure as in the previous section.

\begin{custompromptbox}{Question-Level Aggregation Prompt}
\setlength{\fboxsep}{3pt}  % Reduce padding (default is 6pt)
\small  % Reduce font size slightly
\textcolor{teal}{\textbf{Task:}} \textit{Given the following sub-question answers, provide a concise answer to the original question.}

\textcolor{blue}{\textbf{Original question:}} \texttt{[QUESTION]}

\textcolor{teal}{\textbf{Sub-answers:}}
\begin{itemize}
\item \textit{Facet 1:} \texttt{[FACET\_QUESTION\_1]} → \textit{Answer:} \texttt{[FACET\_ANSWER\_1]}
\item \textit{Facet 2:} \texttt{[FACET\_QUESTION\_2]} → \textit{Answer:} \texttt{[FACET\_ANSWER\_2]}
\item ...
\end{itemize}

\textcolor{purple}{\textbf{Output:}} \textit{Final answer}
\end{custompromptbox}

\section{Model Configuration Details}
\label{app:configuration}

\subsection{Model Configuration Details}

\begin{itemize}
\setlength{\itemsep}{0pt}
\setlength{\parskip}{0pt}
\setlength{\parsep}{0pt}
\setlength{\topsep}{2pt}
\item GPT-4o-mini: Model ID \texttt{gpt-4o-mini-2024-07-18}, \texttt{temperature=0.3}, \texttt{max\_tokens=160} (facet-level), \texttt{max\_tokens=15} (question-level)
\item Gemini-2.0-Flash: Model ID \texttt{models/gemini-2.0-flash}, \texttt{temperature=0.3}, \texttt{max\_output\_tokens=160} (facet-level), \texttt{max\_output\_tokens=15} (question-level)
\item LLaMA-3-8B-Instruct: Via HuggingFace Transformers, 8-bit quantization, \texttt{temperature=0.3}, \texttt{max\_new\_tokens=160} (facet-level), \texttt{max\_new\_tokens=15} (question-level)
\end{itemize}

\subsection{Retrieval Configuration}

\begin{itemize}
\setlength{\itemsep}{0pt}
\setlength{\parskip}{0pt}
\setlength{\parsep}{0pt}
\setlength{\topsep}{2pt}
\item Chunk size: 300 tokens with 50-token overlap. We follow prior empirical findings showing that short, fine-grained chunks improve retrieval effectiveness, consistent with the chunking strategy used by \cite{ramos-varela-etal-2025-context}. \item Top-K: 5 chunks per facet. We set top-k=5 to balance recall and precision: smaller k risks missing relevant evidence, while larger k introduces irrelevant or conflicting context \cite{10.5555/3737916.3741766}. Recent work on RankRAG demonstrates that k=5 works well across diverse QA datasets. 
\item Similarity: Cosine similarity
\end{itemize}

\section{Detailed Experimental Analysis}
\label{app:results}

This appendix provides additional analyses and visualizations that support the main findings reported in Section \ref{main:results}. We focus on distributional behavior, per-question variability, and model-specific patterns that are difficult to convey using aggregate statistics alone.
\begin{figure}[H]
    \centering
    \includegraphics[width=0.45\textwidth,height=0.18\textheight]{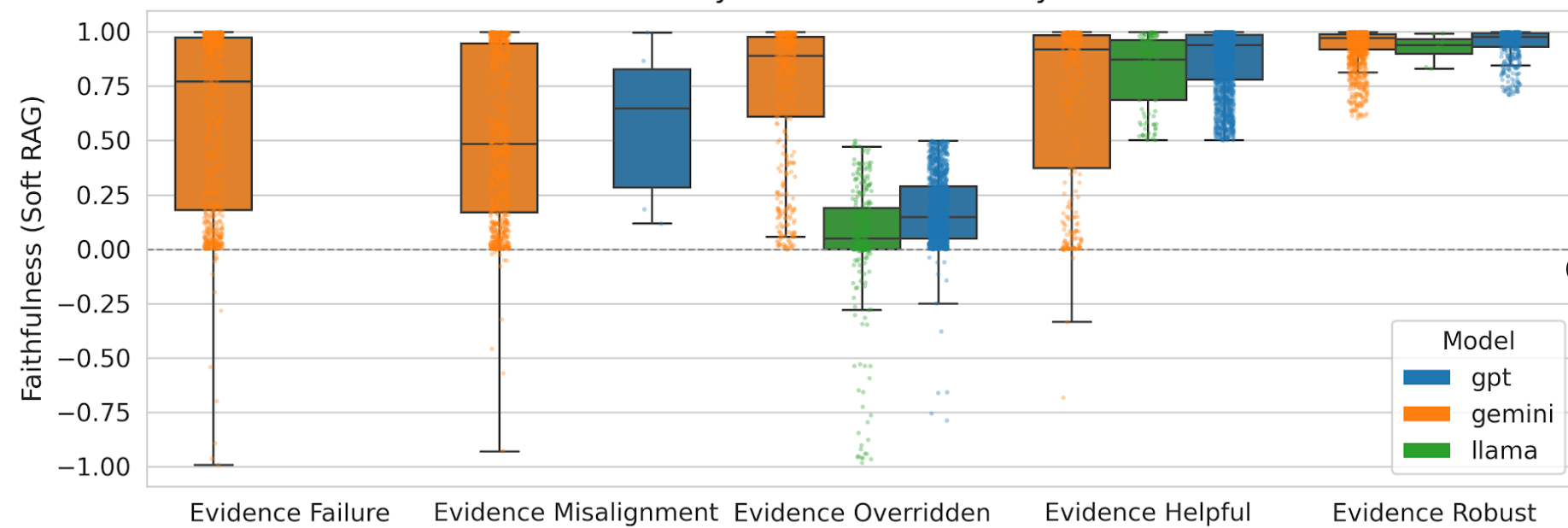}
    \caption{Medical Dataset: Facet-level faithfulness distributions under Soft RAG across evidence taxonomy categories.}
    \label{fig:medical_faithfulness}
\end{figure}
\subsection{Evidence Taxonomy by Question Difficulty}

Figure~\ref{fig:hotpot_difficulty_taxonomy} shows the distribution of evidence taxonomy categories across difficulty levels in HotpotQA. While aggregate results are reported in the main paper, this figure illustrates that Evidence Override remains consistently prevalent across easy, medium, and hard questions, whereas Evidence Robust decreases with increasing difficulty. Evidence Failure and Evidence Misalignment remain relatively rare across all levels, indicating that increased reasoning complexity primarily affects evidence integration during generation rather than retrieval coverage.

\subsection{Facet-Level Faithfulness Distributions}

Figure~\ref{fig:medical_faithfulness} presents the full distribution of facet-level faithfulness scores under Soft RAG for the medical dataset, stratified by evidence taxonomy category. Evidence Helpful and Evidence Robust facets exhibit high faithfulness with low variance across models, while Evidence Failure and Evidence Misalignment show low faithfulness and high variance.
Evidence Overridden facets display a bimodal distribution, with mass concentrated at both high entailment and strong contradiction. This distributional behavior is consistent across datasets, including HotpotQA, and highlights that overridden evidence may be either successfully incorporated or directly contradicted during generation.

\subsection{Per-Question Performance Variability}
\begin{figure}[H]
    \centering
    \includegraphics[width=0.45\textwidth,height=0.18\textheight]{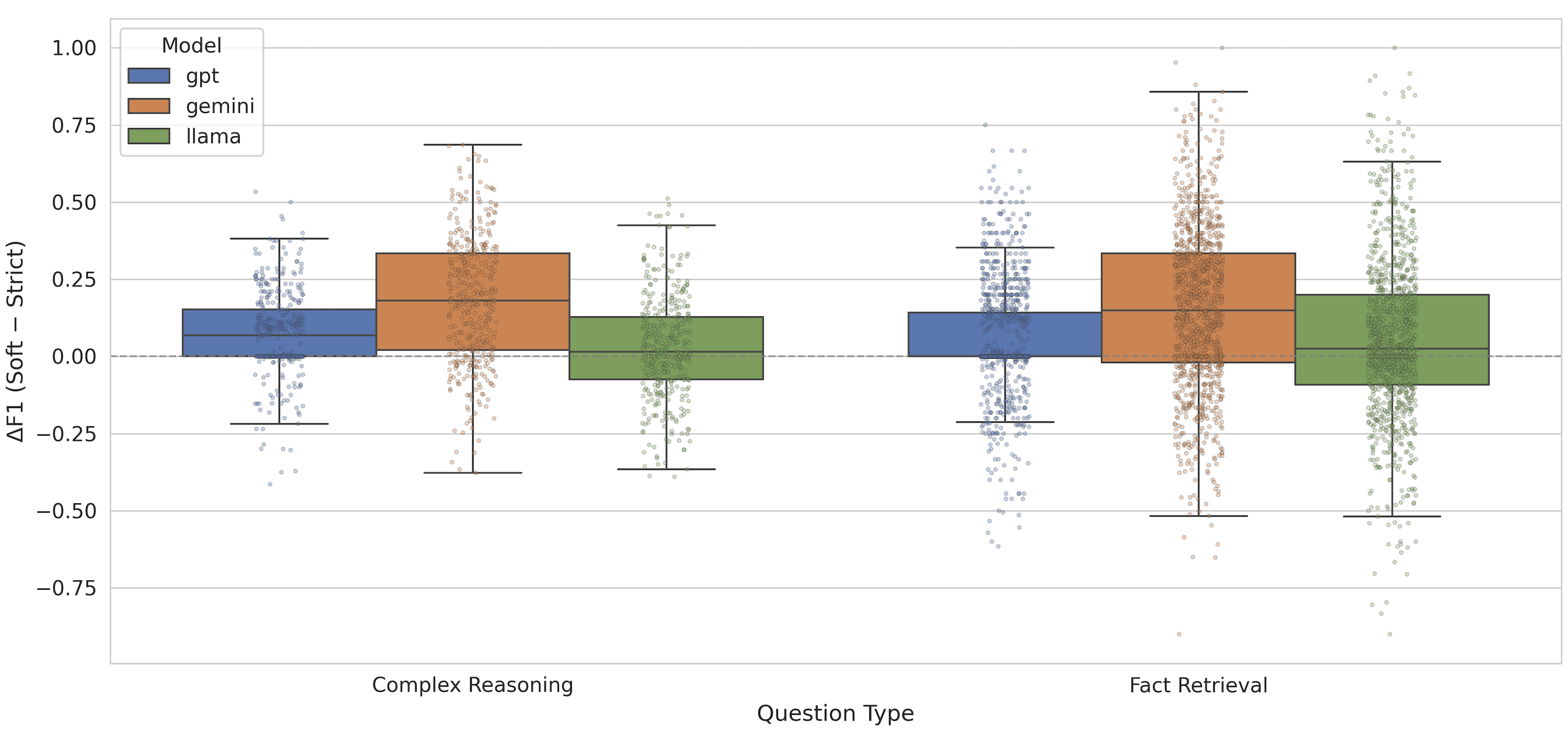}
    \caption{Medical Dataset: Per-question $\Delta$F1 distributions ($\text{Soft} - \text{Strict}$) by model.}
    \label{fig:delta_medical}
\end{figure}
Figure~\ref{fig:delta_medical} visualizes per-question changes in F1 induced by relaxing evidence constraints (Soft RAG vs.\ Strict RAG) for the medical dataset. Although Soft RAG improves average performance, the distribution reveals substantial variability across questions, with a non-trivial fraction experiencing degraded performance. This variability is more pronounced for GPT and LLaMA than for Gemini.

A similar pattern is observed for HotpotQA across difficulty levels, where medium-difficulty questions show the largest gains and hard questions exhibit increased variance. These distributions complement the mean results reported in Section~\ref{main:results} by illustrating the heterogeneity underlying aggregate improvements.

\end{document}